\documentclass[a4paper,12pt]{spieman}  
\usepackage{amsmath,amsfonts,amssymb,bm}
\usepackage{graphicx}
\usepackage{setspace}
\usepackage{tocloft}
\usepackage{epstopdf}
\usepackage{lineno}

\title{Learning-based Natural Geometric Matching with Homography Prior}

\author[a]{Yifang Xu}
\author[a]{Tianli Liao}
\author[a,*]{Jing Chen}
\affil[a]{Nankai University, Center for Combinatorics, 94 Weijin Road, Tianjin, China, 300071}

\cftpagenumbersoff{figure}
\cftpagenumbersoff{table} 
\begin{document} 
\maketitle

\begin{abstract}
Geometric matching is a key step in computer vision tasks. Previous learning-based methods for geometric matching concentrate more on improving alignment quality, while we argue the importance of naturalness issue simultaneously. To deal with this, firstly, Pearson correlation is applied to handle large intra-class variations of features in feature matching stage. Then, we parametrize homography transformation with 9 parameters in full connected layer of our network, to better characterize large viewpoint variations compared with affine transformation. Furthermore, a novel loss function with Gaussian weights guarantees the model accuracy and efficiency in training procedure. Finally, we provide two choices for different purposes in geometric matching. When compositing homography with affine transformation, the alignment accuracy improves and all lines are preserved, which results in a more natural transformed image. When compositing homography with non-rigid thin-plate-spline transformation, the alignment accuracy further improves. Experimental results on Proposal Flow dataset show that our method outperforms state-of-the-art methods, both in terms of alignment accuracy and naturalness.
\end{abstract}


{\noindent \footnotesize\textbf{*}Jing Chen,  \linkable{chenjing@mail.nankai.edu.cn} }

\begin{spacing}{2}   

\section{Introduction}
\label{sect:intro}  
Estimating a geometric matching between images plays a fundamental and important role in computer vision tasks, such as image stitching \cite{Brown:2007,An2015Unified,Li2017Quasi} and image retrieval \cite{Feng2017Spectral}. Traditional geometric matching methods usually start with a local image descriptor and then estimate geometric model by RANSAC \cite{fischler1981random}. In general, there are two main procedures for geometric matching: feature characterization, and transformation model selection. 

\begin{figure*}[]
	\centering
	\includegraphics[width=0.9\textwidth]{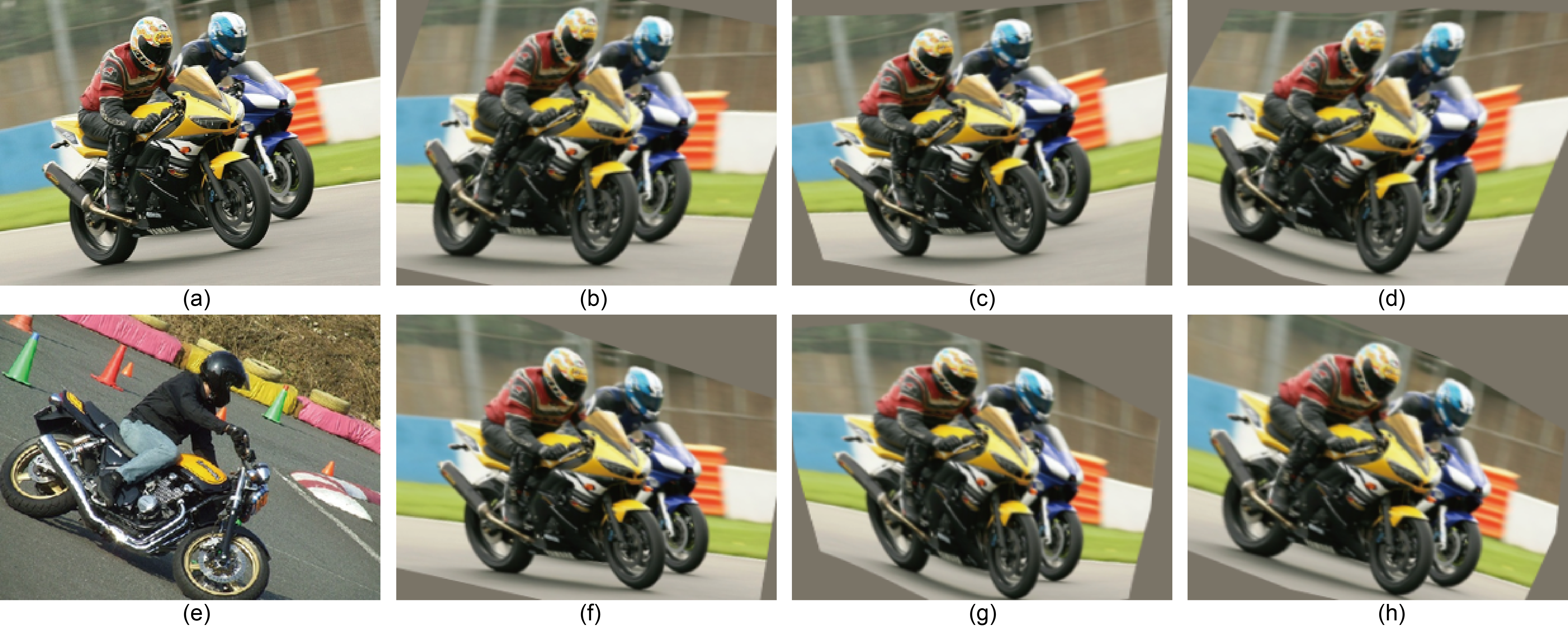}
	\caption{Different transformations on aligning two images. (a) Source image. (b) Affine. (c) TPS. (d) Affine + TPS. (e) Target image. (f) Ours (homography). (g) Ours (homography + TPS). (h) Ours (homography + affine).}
	\label{fig:eg}
\end{figure*}

For feature characterization, many hand-crafted image descriptors \cite{lowe2004distinctive,bay2006surf} were proposed in the last decades, which aimed to reliably localize a set of stable local regions under various imaging conditions. However, when intra-class variation of images is large or background is full of cluster, traditional features may fail to generate valid geometric matching.
\begin{figure}[!t]
	\centering
	\includegraphics[width=0.9\textwidth]{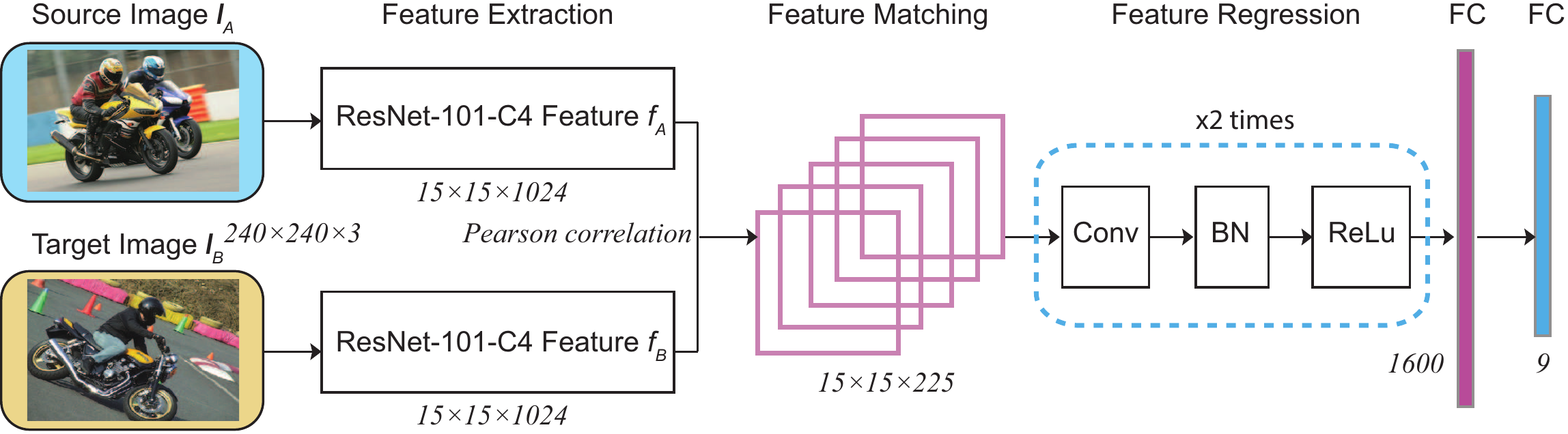}
	\caption{Architecture of our geometric matching network, where we keep the feature regression stage same as \cite{Rocco2017Convolutional}.}
	\label{fig:arch}
\end{figure}
For transformation model selection, global $2$D transformations such as Euclidean, similarity, affine, and homography have occupied a key position in rigid transformations. Among these transformations, affine has 6 degrees of freedom which maps parallel lines to parallel lines but lacks power of characterizing large viewpoint variations, whereas homography has 8 degrees of freedom which has better alignment, preserve all straight lines, and characterizes viewpoint variations well. To achieve alignment flexibility and accuracy, thin-plate spline (TPS) \cite{bookstein1989principal} and as-projective-as-possible (APAP) warp \cite{zaragoza2014projective}, formulate the non-rigid transformation as a matching problem with a smoothness constraint. However, naturalness is also a critical factor which influences the appearance of final transformed image. Non-rigid transformations usually align the source image and target image better than rigid transformations, but they suffer from line-bending and shape-distortion to some extent.  As shown in Fig. \ref{fig:eg}(c), TPS transformation aligns the source image and target image well, at the cost of local distortion (such as curved lines).

Recently, learning-based geometric matching methods \cite{han2015matchnet,simo2015discriminative,detone2016deep,Rocco2017Convolutional} have made a great progress on dealing with the limited generalization power of traditional feature descriptors and the low alignment accuracy under large intra-class variations. Unlike \cite{han2015matchnet,simo2015discriminative,detone2016deep} which divide the image into
local patches and extract descriptors individually, Rocco et al. \cite{Rocco2017Convolutional} proposed an end-to-end convolutional neural network architecture, which combines affine and TPS transformations (see Fig. \ref{fig:eg}(d)) for coarse-to-fine alignment. However, affine transformation is not flexible enough to characterize large viewpoint variations. As shown in Fig. \ref{fig:eg}(b), the orientation of motorbike after affine transformation looks different from that in target image, whereas homography can characterize this orientation properly (see Fig. \ref{fig:eg}(f)).

The above alignment and naturalness issues motivate us to construct a geometric matching which aligns images well and produces natural transformed images. Since affine is not powerful for aligning images and characterizing large viewpoint variations, we use homography in our network. The composition of homography and  TPS provides better alignment (see Fig. \ref{fig:eg}(g)) than state-of-the-art methods, while the composition of homography and affine results in a more natural transformed image which preserves all lines (see Fig. \ref{fig:eg}(h)). In this letter, we propose a novel learning-based architecture, which can be trained end-to-end for natural geometric matching with homography prior.
Contributions are summarized as follows:
\begin{enumerate}
	\item We introduce Pearson correlation for improving the generalization ability of feature matching stage.
	\item We add homography transformation to our CNN architecture for learning more flexible rigid transformation.
	\item We define a new loss function for training the model, which equips a location dependent Gaussian weight for grid loss \cite{Rocco2017Convolutional}.
	\item Our single homography model achieves the best performance on Proposal Flow dataset, while the compositions of homography and other transformations solve alignment and naturalness issues successfully.
\end{enumerate}

\section{Proposed Architecture}
In this section, a novel architecture (see Fig. \ref{fig:arch}) is proposed for natural geometric matching. After extracting the features by CNN, we use Pearson correlation \cite{Pearson2006Note} to match features of two input images. Then, we apply the homography transformation with more degrees of freedom to obtain better alignment and more natural transformed images. Finally, a novel loss function guarantees the model accuracy and efficiency in training procedure. Compositing homography with affine or TPS transformation significantly further increases the performance. 

\subsection{Feature Extraction}
Given two input images $I_A$ and $I_B$ with size $h\times w\time3$ ($240\times240$ in our architecture), we use ResNet-101 \cite{he2016deep} cropped at the \texttt{layer3} layer (ResNet-101-C4 layer in \cite{he2016deep}) to extract deep features in two input images, respectively. Compared to VGG-16 network \cite{simonyan2015very}, ResNet-101 has been proven that it can characterize the features better and achieve less error on a majority of datasets.
\subsection{Matching with Pearson Correlation}
In \cite{Rocco2017Convolutional}, given L2-normalized dense feature
maps $f_A$, $f_B\in \mathbb{R}^{h\times w\times d}$, the correlation map of between features is then characterized by scalar product of individual column vectors. Essentially, let $\bm{x}\in\mathbb{R}^d$ and $\bm{y}\in\mathbb{R}^d$ be the column vectors in $f_A$ and $f_B$, the above two-step feature normalization and matching procedures can be regard as cosine similarity,
\begin{equation}
cos(\bm{x},\bm{y})=\frac{\bm{x}\bm{\cdot}\bm{y}}{\|\bm{x}\|\|\bm{y}\|}.
\end{equation}
However, cosine similarity is not robust to shifts. If the distribution of $\bm{x}$ or $\bm{y}$ has large intra-class variations, the cosine similarity would change sensitively at the same time, which greatly effects the generalization power of feature matching stage.

Fortunately, in statistics, Pearson correlation \cite{Pearson2006Note} can measure the  correlation or similarity between two variables $\bm{x}$ and $\bm{y}$ well. Thus, in our feature matching stage, we replace original correlation map $c_{AB}\in\mathbb{R}^{h\times w\times(h\times w)}$ of individual descriptor $\bm{f}_A\in f_A$ and $\bm{f}_B\in f_B$ in \cite{Rocco2017Convolutional} with Pearson correlation, i.e.,
\begin{equation}
c_{AB}(i,j,k)=\frac{(\bm{f}_B(i,j)-\bar{\bm{f}}_B)^T(\bm{f}_A(i_k,j_k)-\bar{\bm{f}}_A)}{\|\bm{f}_B(i,j)-\bar{\bm{f}}_B\|\|\bm{f}_A(i_k,j_k)-\bar{\bm{f}}_A\|},
\end{equation}
where $k=h(j_k-1)+i_k$, $\bar{\bm{f}_A}$ and $\bar{\bm{f}_B}$ are the respective means. The new correlation map $c_{AB}$ alleviates the influence of large intra-class variations, and then makes our matching stage more general and robust under different conditions.
\subsection{Transformation Model Selection}
Estimating a geometric transformation is a challenging task in deep learning in recent years, since selecting a proper transformation is not intuitive and always needs more considerations about variations of different scenes. As mentioned above, considering the disadvantages of affine and TPS, we use homography transformation, which is known as the most flexible rigid transformation, in our geometric matching architecture for better alignment and more natural transformed images.

Generally, we can parametrize homography to be $8$ parameters, which contains a $3\times3$ matrix $\bm{H}$ with a fixed scale constraint \cite{detone2016deep}. As pointed in \cite{hartley2003multiple}, it is not necessary and advisable to use minimal parametrization since the surface of the cost function usually becomes more complicated when minimal parametrizations are used. Thus, we parametrize homography transformation with $9$ parameters, i.e, a $3\times3$ matrix with the following form
\begin{equation}
\bm{H}=\left(
\begin{array}{ccc}
h_{1} & h_{2} & h_{3}\\
h_{4} & h_{5} & h_{6} \\
h_{7} & h_{8} & h_{9}
\end{array}\right),
\end{equation}
and allow the last parameter $h_9$ to be an arbitrary real number. 

Compared to affine and TPS transformations, homography has a better alignment ability and viewpoint characterization than affine, and can preserve all lines while TPS always fails. After compositing homography with affine and TPS transformations in \cite{Rocco2017Convolutional}, we have two choices: affine + homography, and homography + TPS. Compared with affine + TPS in \cite{Rocco2017Convolutional}, the first choice results in a more natural transformed image (see Fig. \ref{fig:eg}), and the second choice achieves better alignment accuracy (see Tab. \ref{tab:comp}).
\subsection{Loss Function}
Let $\mathcal{T}_{\theta_{GT}}$ and $\mathcal{T}_{\hat{\theta}}$ denote the ground truth and the estimated transformations with parameters $\theta_{GT}$ and $\hat{\theta}$, respectively. Rocco et al. \cite{Rocco2017Convolutional} proposed a grid loss which measures the discrepancy between the two transformed imaginary grids by
\begin{equation}
\mathcal{L}_{grid}(\hat{\theta},\theta_{GT})=\frac{1}{N}\sum_{i=1}^Nd(\mathcal{T}_{\hat{\theta}}(g_i),\mathcal{T}_{\theta_{GT}}(g_i))^2,
\end{equation}
where $\{g_i\}=\{(x_i,y_i)\}$ is the set of grid point from $-1$ to $1$ with the step $0.1$, and $N$ is the number of grid points, i.e., $N=20\times20$. 

\begin{figure}[!t]
	\centering
	\includegraphics[width=0.7\textwidth]{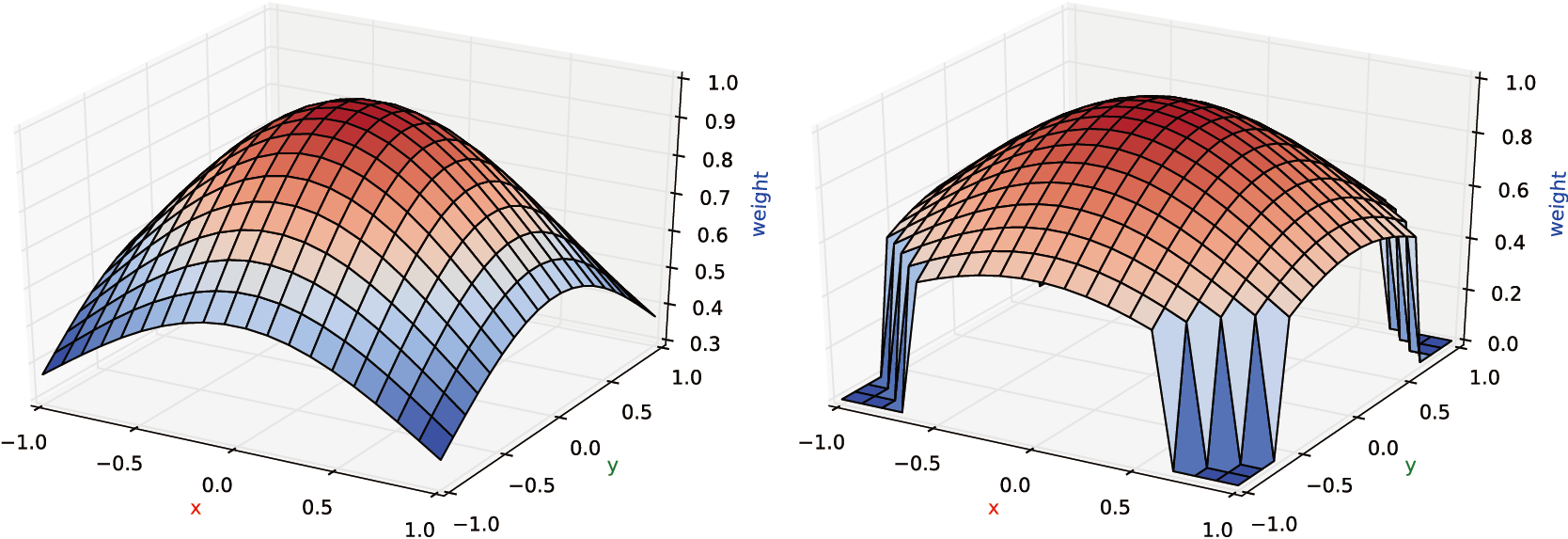}
	\caption{Gaussian weight w/o the parameter $\gamma$ in our loss function.}
	\label{fig:weight}
\end{figure}

We mimic the classical approach APAP \cite{zaragoza2014projective} which uses Gaussian weights to give higher importance to data
that are closer to features for better alignment. Since objects always locate in the center of the image, we give higher weights to  respects the local structure around the center $(x_c,y_c)$ of grid points (see Fig. \ref{fig:weight}), and the weight in grid $g_i$ is calculated as
\begin{equation}
\omega_i = \begin{cases}
e^{-\frac{(x_i-x_c)^2+(y_i-y_c)^2}{2\sigma^2}}, \text{if $e^{-\frac{(x_i-x_c)^2+(y_i-y_c)^2}{2\sigma^2}}\ge \gamma$}.\\
0, \text{if $e^{-\frac{(x_i-x_c)^2+(y_i-y_c)^2}{2\sigma^2}}<\gamma$}
\end{cases}
\end{equation}
Thus, our loss is defined as
\begin{equation}
\mathcal{L}(\hat{\theta},\theta_{GT})= \sum_{i=1}^N\omega_id(\mathcal{T}_{\hat{\theta}}(g_i),\mathcal{T}_{\theta_{GT}}(g_i))^2. \label{loss}
\end{equation}
Since $\mathcal{L}_{grid}$ is trainable with respect to the parameters of homography, our loss $\mathcal{L}$ is differentiable with respect to $\hat{\theta}$ in a straightforward manner.

\section{Experimental Evaluation}
In this section, we compare proposed architecture against CNN geometric matching methods \cite{Rocco2017Convolutional} and several state-of-the-art geometric matching methods \cite{duchenne2011graph,Liu2011SIFT,Kim2013Deformable,revaud2016deepmatching,Ham2016Proposal} on Proposal Flow dataset \cite{Ham2016Proposal}. Moreover, ablation studies further show the advantages of proposed method.

In our experiment, we set hyper-parameters following existing architecture  as \cite{Rocco2017Convolutional}, set $\sigma$ to be $1$, and $\gamma$ to be $0.5$ in Eq. \eqref{loss} since the grids range from $[-1,1]$. Codes are implemented in PyTorch \cite{paszke2017pytorch}, and based on the code \footnote{\url{http://github.com/ignacio-rocco/cnngeometric_pytorch}} of \cite{Rocco2017Convolutional}.
\begin{table}[!t]
	\centering
	\caption{Performance comparisons on Proposal Flow dataset \cite{Ham2016Proposal}.}\label{tab:comp}
	\begin{tabular}{lc}
		\hline
		Methods           &   PCK (\%)  \\
		\hline \hline
		DeepFlow \cite{revaud2016deepmatching}  & 20          \\
		GMK \cite{duchenne2011graph}       & 27       \\
		SIFT Flow \cite{Liu2011SIFT} & 38       \\
		DSP \cite{Kim2013Deformable}       & 29       \\
		Proposal Flow NAM \cite{Ham2016Proposal} & 53  \\
		Proposal Flow PHM \cite{Ham2016Proposal} & 55 \\
		Proposal Flow LOM \cite{Ham2016Proposal} & 56 \\
		VGG + CNNGeo. (affine) \cite{Rocco2017Convolutional} & 49 \\
		ResNet-101 + CNNGeo. (affine) \cite{Rocco2017Convolutional} & 56 \\
		ResNet-101 + CNNGeo. (TPS) \cite{Rocco2017Convolutional} & 58 \\
		ResNet-101 + Ours (homo) & \textbf{61}\\
		\hline \hline	
		VGG + CNNGeo. (affine + TPS) \cite{Rocco2017Convolutional} & 56 \\
		ResNet-101 + CNNGeo. (affine + TPS) \cite{Rocco2017Convolutional} & 68 \\
		ResNet-101 + Ours (homo + affine) & \textbf{65}\\
		ResNet-101 + Ours (homo + TPS) & \textbf{69}\\
		
		\hline
	\end{tabular}
\end{table}
\begin{table}[!t]
	\centering
	\caption{Ablation studies for single homogrpahy.}\label{tab:Ablation}
	\begin{tabular}{lc}
		\hline
		Methods           &   Pascal-synth-homo  \\
		\hline \hline
		Matching with cosine similarity  \cite{Rocco2017Convolutional}  & 58\\
		Homography with $8$ parameters & 55\\
		Training with MSE loss & 55\\
		Training with grid loss & 59\\
		Ours with our loss + Pearson correlation + $9$ parameters & \textbf{61}\\
		
		\hline
	\end{tabular}
\end{table}
\begin{figure*}[!t]
	\centering
	\includegraphics[width=0.95\textwidth]{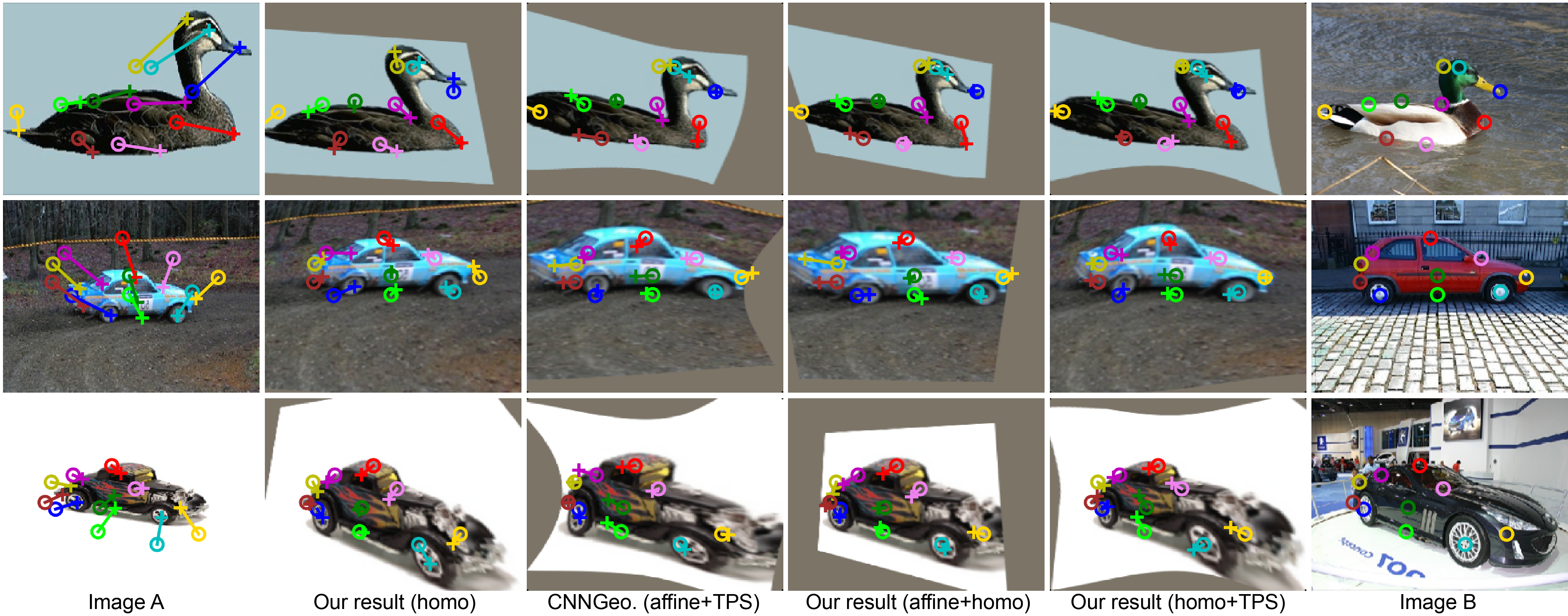}
	\caption{ Qualitative comparisons on the Proposal Flow dataset. Our method handles scale, rotation, translation, and viewpoint variations successfully.}
	\label{fig:vis}
\end{figure*}
\subsection{Datasets and Evaluation Criterion
}
We train the proposed network for homography on Pascal VOC 2011 by strongly supervised learning, with the generated dataset \emph{Pacal-synth-homo}. For each image in the dataset, we randomly perturb the four corners of the image within $1/4$ image, then computed a relative homography matrix by the coordinates of four corners. Thus, each data contains a source image, a homography matrix with $9$ numbers, and a transformed image by homography. To evaluate the proposed architecture, we follow the standard criterion, the average probability of correct keypoint (PCK) \cite{yang2013articulated}, which computes the proportion of keypoints that are correctly matched. 
\subsection{Comparisons with state-of-the-arts}
We run a number of comparisons to analyze the proposed architecture. As shown in Tab. \ref{tab:comp}, our architecture with single homography reaches the first place in all single model. When compositing with TPS, our architecture achieves better performance than the competing methods in \cite{Rocco2017Convolutional} which composites affine and TPS. Note that the ResNet-101-based model in \cite{Rocco2017Convolutional} uses larger datasets, and is composed of many other strategies, while our model does not use them but still performs better. Although compositing our homography with affine does not get the best performance, it is still meaningful when considering the naturalness issue, especially when the scenes is full of line structures. 

Figure \ref{fig:vis} shows a qualitative comparison of geometric matching on Proposal Flow dataset. Compared with \cite{Rocco2017Convolutional}, our homography + TPS has less alignment error (see the first and the second examples), and homography + affine produces a more natural result (see the third example).

\subsection{Ablation Studies}
Extensive ablation studies are conducted to validate how each of these changes contribute to our overall architecture, as shown in Tab. \ref{tab:Ablation}. When matching with cosine similarity as \cite{Rocco2017Convolutional}, the performance drops $3$ percent. After parametrizing homography with $9$ degrees of freedom instead of $8$, the PCK increases significantly. Moreover, our loss function effectively improves the performance compared to grid loss and MSE loss.

\section{Conclusion}
We propose a novel architecture for natural geometric matching with homography prior, which uses Pearson correlation to match features of two input images, parametrizes the homography transformation with more degrees of freedom, and applies a novel loss function to guarantee the model accuracy and efficiency in training procedure. Compositing our homography prior with affine or TPS transformation significantly increases the performance both on alignment accuracy and naturalness. Future works include making the separate training of homography and other transformations in a unified framework.


\bibliographystyle{ieeeabbr}   

\end{spacing}
\end{document}